\def\year{2020}\relax
\documentclass[letterpaper]{article} % DO NOT CHANGE THIS
\usepackage{aaai20}  % DO NOT CHANGE THIS
\usepackage{times}  % DO NOT CHANGE THIS
\usepackage{helvet} % DO NOT CHANGE THIS
\usepackage{courier}  % DO NOT CHANGE THIS
\usepackage[hyphens]{url}  % DO NOT CHANGE THIS
\usepackage{graphicx} % DO NOT CHANGE THIS
\usepackage{graphicx}  % DO NOT CHANGE THIS
\usepackage{amsmath}
\usepackage{amssymb}
\usepackage{algorithm,algorithmic}
\title{Learning Relationships between Text, Audio, and Video via Deep Canonical Correlation for Multimodal Language Analysis}

\author{
Zhongkai Sun,\textsuperscript{\rm 1}
Prathusha K Sarma,\textsuperscript{\rm 2}\thanks{Work done while at UW-Madison}
William A Sethares,\textsuperscript{\rm 1}
Yingyu Liang\textsuperscript{\rm 1}\\ 
\textsuperscript{\rm 1}University of Wisconsin-Madison, 
\textsuperscript{\rm 2}Curai\\ 
zsun227@wisc.edu, prathyushaks.21@gmail.com, sethares@wisc.edu, yliang@cs.wisc.edu
}
\begin{document}

\maketitle

\begin{abstract}
Multimodal language analysis often considers relationships between features based on text and those based on acoustical and visual properties. Text features typically outperform non-text features in sentiment analysis or emotion recognition tasks in part because the text features are derived from advanced language models or word embeddings trained on massive data sources while audio and video features are human-engineered and comparatively underdeveloped. Given that the text, audio, and video are describing the same utterance in different ways, we hypothesize that the multimodal sentiment analysis and emotion recognition can be improved by learning (hidden) correlations between features extracted from the outer product of text and audio (we call this {\em text-based audio}) and analogous {\em text-based video}. This paper proposes a novel model, the Interaction Canonical Correlation Network (ICCN), to learn such multimodal embeddings. ICCN learns correlations between all three modes via deep canonical correlation analysis (DCCA) and the proposed embeddings are then tested on several benchmark datasets and against other state-of-the-art multimodal embedding algorithms. Empirical results and ablation studies confirm the effectiveness of ICCN in capturing useful information from all three views. 
\end{abstract}
\section{Introduction}
Human language communication occurs in several modalities: via words that are spoken, by tone of voice, and by facial and bodily expressions. Understanding the content of a message thus requires understanding all three modes. With the explosive growth in availability of data, several machine learning algorithms have been successfully applied towards multimodal tasks such as sentiment analysis \cite{morency_towards_2011,soleymani2017survey}, emotion recognition \cite{haq2011multimodal}, image-text retrieval \cite{wang2016learning}, and aiding medical diagnose \cite{liu2019,lee2014supervised} etc. 
Among multimodal language sentiment or emotion experiments involving unimodal features~\cite{zadeh2016mosi,zadeh2018multimodal,tsai2018learning,tsai2019multimodal}, it is commonly observed that text based features perform better than visual or auditory modes. This is plausible for at least three reasons: 1) Text itself contains considerable sentiment-related information. 2) Visual or acoustic information may sometimes confuse the sentiment or emotion analysis task. For instance: ``angry'' and ``excited'' may have similar acoustic performances (high volume and high pitch) even though they belong to opposite sentiments. Similarly, ``sad'' and ``disgusted'' may have different visual features though they both belong to the negative sentiment. 3) Algorithms for text analysis have a richer history and are well studied.

Based on this observation, learning the hidden relationship between verbal information and non-verbal information is a key point in multi-modal language analysis. This can be approached by looking at different ways of combining multi-modal features.

The simplest way to combine text (T), audio (A) and video (V) for feature extraction and classification is to concatenate the A, V, and T vectors. An alternative is to use the outer product \cite{liu2018efficient,zadeh2017tensor} which can represent the interaction between pairs of features, resulting in 2D or 3D arrays that can be processed using advanced methods such as convolutional neural networks (CNNs) \cite{lawrence1997face}. Other approaches \cite{zadeh2018memory,liang2018multimodal,zadeh2018multimodal,wang2018words} study multi-modal interactions and intra-actions by using either graph or temporal memory networks with a sequential neural network LSTM \cite{gers1999learning}. While all these have contributed towards learning multi-modal features, they typically ignore the hidden correlation between text-based audio and text-based video. Individual modalities are either combined via neural networks or passed directly to the final classifier stage. However, it is obvious that attaching both audio and video features to the same textual information may enable non-text information to be better understood, and in turn the non-text information may impart greater meaning to the text. Thus, it is reasonable to study the deeper correlations between text-based audio and text-based video features.

This paper proposes a novel model which uses the outer-product of feature pairs along with Deep Canonical Correlation Analysis (DCCA) \cite{andrew2013deep} to study useful multi-modal embedding features. The effectiveness of using an outer-product to extract cross-modal information has been explored in \cite{zadeh2017tensor,liu2018efficient}. Thus, features from each mode are first extracted independently at the sentence (or utterance) level and two outer-product matrices ($T \otimes V$ and $T \otimes A$) are built for representing the interactions between text-video and between text-audio. Each outer-product matrix is connected to a convolutional neural network (CNN) for feature extraction. Outputs of these two CNNs can be considered as feature vectors for text-based audio and text-based video and should be correlated.

In order to better correlate the above text-based audio and text-based video, we use Canonical Correlation Analysis (CCA) \cite{10.1093/biomet/28.3-4.321}, which is a well-known method for finding a linear subspace where two inputs are maximally correlated. Unlike cosine similarity or Euclidean distance, CCA is able to learn the direction of maximum correlation over all possible linear transformations and is not limited by the original coordinate systems. However, one limitation of CCA is that it can only learn linear transformations. An extension to CCA named Deep CCA (DCCA) \cite{andrew2013deep} uses a deep neural network to allow non-linear relationships in the CCA transformation. Recently several authors \cite{rotman2018bridging,hazarika2018cascade} have shown the advantage of using CCA-based methods for studying correlations between different inputs. Inspired by these, we use DCCA to correlate text-based audio and text-based video. Text-based audio and text-based video features derived from the two CNNs are input into a CCA layer which consists of two projections and a CCA Loss calculator. The two CNNs and the CCA layer then form a DCCA, the weights of the two CNNs and the projections are updated by minimizing the CCA Loss. In this way, the two CNNs are able to extract useful features from the outer-product matrices constrained by the CCA loss. After optimizing the whole network, outputs of the two CNNs are concatenated with the original text sentence embedding as the final multi-modal embedding, which can be used for the classification.

We evaluate our approach on three benchmark multi-modal sentiment analysis and emotion recognition datasets: CMU-MOSI \cite{zadeh2016mosi}, CMU-MOSEI \cite{zadeh2018multimodal}, and IEMOCAP\cite{busso2008iemocap}. Additional experiments are presented to illustrate the performance of the ICCN algorithm. 
The rest of the paper is organized as follows: Section~\ref{relwork} presents related work, Section~\ref{methods} introduces our proposed model and Section~\ref{exps} describes our experimental setup. Section~\ref{discu} presents a discussion on the empirical observations, Section~\ref{conclu} concludes this work.

\section{Related Work}\label{relwork}
The central themes of this paper are related to learning (i) multi-modal fusion embeddings and (ii) cross-modal relationships via canonical correlation analysis (CCA).

\noindent

\textbf{Multi-modal fusion embedding:} Early work~\cite{poria2016convolutional} concatenates the audio, video and text embeddings to learn a larger multi-modal embedding. But this may lead to a potential loss of information between different modalities. Recent studies on learning multi-modal fusion embeddings train specific neural network architectures to combine all three modalities. In their work~\cite{chen2017multimodal} propose improvements to multi-modal embeddings using reinforcement learning to align the multi-modal embedding at the word level by removing noises. A multi-modal tensor fusion network is built in~\cite{zadeh2017tensor} by calculating the outer-product of text, audio and video features to represent comprehensive features. However this method is limited by the need of a large computational resources to perform calculations of the outer dot product. In their work~\cite{liu2018efficient} developed an efficient low rank method for building tensor networks which reduce computational complexity and are able to achieve competitive results. A Memory Fusion Network (MFN) is proposed by~\cite{zadeh2018memory} which memorizes temporal and long-term interactions and intra-actions between cross-modals, this memory can be stored and updated in a LSTM.~\cite{liang2018multimodal} learned multistage fusion at each LSTM step so that the multi-modal fusion can be decomposed into several sub-problems and then solved in a specialized and effective way. A multimodal transformer is proposed by~\cite{tsai2019multimodal} that uses attention based cross-modal transformers to learn interactions between modalities.

\noindent

\textbf{Cross-modal relationship learning via CCA:} Canonical Correlation Analysis(CCA)~\cite{10.1093/biomet/28.3-4.321} learns the maximum correlation between two variables by mapping them into a new subspace. Deep CCA (DCCA)~\cite{andrew2013deep} improves the performance of CCA by using feedforward neural networks in place of the linear transformation in CCA. 

A survey of recent literature sees applications of CCA-based methods in analyzing the potential relationship between different variables. For example, a CCA based model to combine domain knowledge and universal word embeddings is proposed by~\cite{sarma2018domain}. Models proposed by~\cite{rotman2018bridging} use Deep Partial Canonical Correlation Analysis (DPCCA), a variant of DCCA, for studying the relationship between two languages based on the same image they are describing. Work by~\cite{sun2019multi} investigates the application of DCCA to simple concatenations of multimodal-features, while~\cite{hazarika2018cascade} applied CCA methods to learn joint-representation for detecting sarcasm. Both approaches show the effectiveness of CCA methods towards learning potential correlation between two input variables.
\section{Methodology}\label{methods}

This section first introduces CCA and DCCA. Next, the interaction canonical correlation network (ICCN), which extracts the interaction features of a CNN in a DCCA-based network, is introduced. Finally, the whole pipe-line of using this method in a multimodal language analysis task is described.

\subsection{CCA and DCCA}

Given two sets of vectors  $X \in \mathbb{R}^{n_{1} \times m}$ and $Y \in {\mathbb{R}}^{n_{2}\times m}$, where $m$ denotes the number of vectors, CCA learns two linear transformations $A \in \mathbb{R}^{n_{1} \times r}$ and $B \in \mathbb{R}^{n_{2} \times r}$ such that the correlation between $A^{T}X$ and $B^{T}Y$ is maximized. Denote the covariances of $X$ and $Y$ as $S_{11}, S_{22}$, and the cross-covariance of $X, Y$ as $S_{12}$. The CCA objective is  
\begin{equation}
    \begin{split}
        A^{*}, B^{*} &=  \mathop{\arg\max}_{A, B}  \textrm{corr}(A^{T}X, B^{T}Y)\\
        &= \mathop{\arg\max}_{A, B} \frac{A^{T}S_{12}B}{\sqrt{A^{T}S_{11}A} \sqrt{B^{T}S_{22}B}}.
    \end{split}
\end{equation}
The solution of the above equation is fixed and can be solved in multiple ways \cite{10.1093/biomet/28.3-4.321,martin1979multivariate}. One method suggested by \cite{martin1979multivariate} lets $U,S,V^{T}$ be the Singular Value Decomposition ($SVD$) of the matrix $Z$ = $S_{11}^{-\frac{1}{2}}S_{12}S_{22}^{-\frac{1}{2}}$. Then $A^{*}, B^{*}$ and the total maximum canonical correlation are
\begin{equation} \label{eqn:maxCCA}
    \begin{split}
        A^{*} &= S_{11}^{-\frac{1}{2}}U\\
        B^{*} &= S_{22}^{-\frac{1}{2}}V\\
        \textrm{corr}(A^{*T}X, B^{*T}Y) &= \textrm{trace}(Z^{T}Z)^{\frac{1}{2}}.
    \end{split}
\end{equation}
One limitation of CCA is that it only considers linear transformations. DCCA \cite{andrew2013deep} learns non-linear transformations using a pair of neural networks. Let $f, g$ denote two independent neural networks, the objective of DCCA is to optimize parameters $\theta_{f}, \theta_{g}$ of $f, g$ so that the canonical correlation between the output of $f $ and $ g $, denoted as $F_{X} = $ $f(X;\theta_{1})$ and $F_{Y} =$ $g(Y;\theta_{2})$, can be maximized by finding two linear transformations $C^{*}, D^{*}$. The objective of DCCA is
\begin{equation}
    \begin{split}
            \theta_{f}^{*}, \theta_{g}^{*} &=  \mathop{\arg\max}_{\theta_{f}, \theta_{g}}  \textrm{CCA}(F_{X}, F_{Y})\\
            &= \mathop{\arg\max}_{\theta_{f}, \theta_{g}} \textrm{corr}(C^{*T}F_{X}, D^{*T}F_{Y}) .
    \end{split}
\end{equation}
In order to update the parameters of $f, g$, a loss for measuring the canonical correlation must be calculated and back-propagated. Let $R_{11}, R_{22}$ be covariances of $F_{X}, F_{Y}$, the cross-covariance of $F_{X}, F_{Y}$ as $R_{12}$. Let $E = R_{11}^{-\frac{1}{2}}R_{12}R_{22}^{-\frac{1}{2}}$. 
According to \eqref{eqn:maxCCA}, the canonical correlation loss for updating $F_{x} , F_{Y}$ can be defined by
\begin{equation} \label{eqn:CCALoss}
    \textrm{CCA Loss}  = - \textrm{trace}(E^{T}E)^{\frac{1}{2}}.
\end{equation}
Networks $f(X;\theta_{f}), g(Y;\theta_{g})$'s parameters can be updated by minimizing the CCA Loss \eqref{eqn:CCALoss} (i.e. maximize the total canonical correlation).

\subsection{Text Based Audio Video Interaction Canonical Correlation}
Previous work of~\cite{zadeh2017tensor,liu2018efficient} on multi-modal feature fusion has shown that the outer-product is able to learn interactions between different features effectively. Thus, we use the outer-product to represent text-video and text-audio features. Given that  outer-product and DCCA are applied at the utterance (sentence)-level, we extract utterance level features for each uni-modal independently in order to test the effectiveness of ICCN more directly. Let $H_{t} \in \mathbb{R}^{d_{t}}$ be the utterance-level text feature embedding, and $H_{v} \in \mathbb{R}^{d_{v}\times l_{v}}, H_{a} \in \mathbb{R}^{d_{a}\times l_{a}}$ be the video and audio input sequences. A 1D temporary convolutional layer is used to extract local structure of the audio and video sequences, and the outputs of the 1D-CNN are denoted as $H_{a1} \in \mathbb{R}^{d_{a1}\times l_{a}}, H_{v1} \in \mathbb{R}^{d_{v1}\times l_{v}}$. Next, two LSTMs process the audio and video sequences. The final hidden state of each LSTM is used as the utterance-level audio or video feature, denoted as $H_{a2}\in \mathbb{R}^{d_{a2}}, H_{v2}\in \mathbb{R}^{d_{v2}}$. Once each utterance-level feature has been obtained, the text-based audio feature matrix and text-based video feature matrix can then be learned using the outer-product on $H_{t} , H_{v2}, H_{a2}$: 

\begin{equation}
    \begin{split}
        H_{ta} &= H_{t} \otimes H_{a2}, H_{ta} \in \mathbb{R}^{d_{t}\times d_{a2}} \\
        H_{tv} &= H_{t} \otimes H_{v2}, H_{tv} \in \mathbb{R}^{d_{t}\times d_{v2}}.
    \end{split}
\end{equation}
In order to extract useful features from the outer-product matrices $H_{ta}, H_{tv}$, a Convolutional Neural Network is used as the basic feature extractor.  $H_{ta}$ and $H_{tv}$ are connected by multiple 2D-CNN layers with max pooling. Outputs of the two 2D-CNNs are reshaped to 1D vector and then be used as inputs to the CCA Loss calculation. 1D-CNN, LSTM, and 2D-CNN's weights are again updated using the back-propagation of the CCA Loss~\eqref{eqn:CCALoss}. Thus the two 2D-CNNs learn to extract features from $H_{tv}$ and $H_{ta}$ so that their canonical correlation is maximized.
\begin{figure*}[ht!]
    \centering
    \includegraphics[width=1.75\columnwidth]{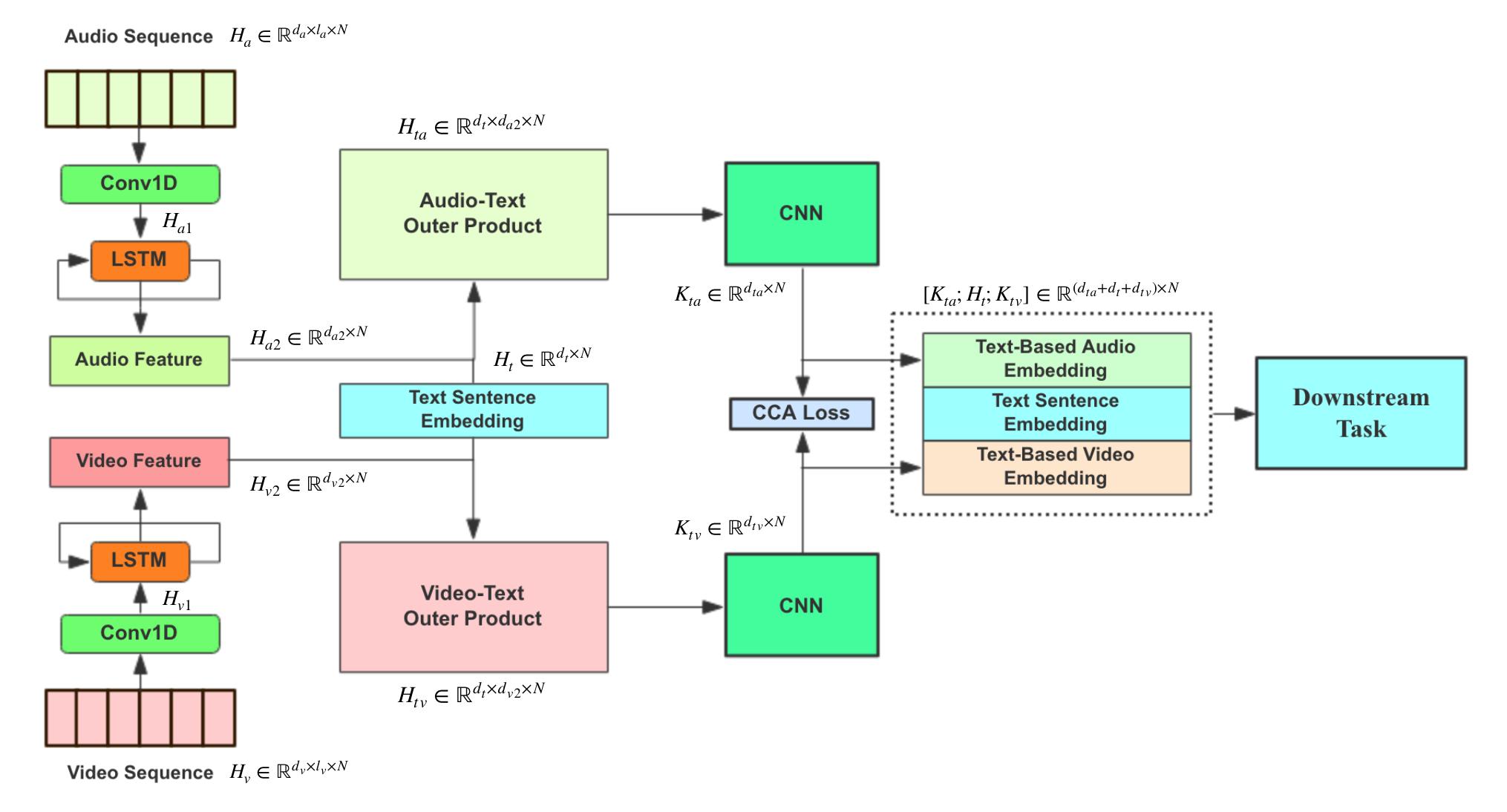}
    \caption{ICCN method for aligning text-based audio features and text-based video features. Sentence level uni-modal features are extracted independently. Outer-product matrices of text-audio and text-video are used as input to the Deep CCA network. After learning the CNN's weights using the CCA Loss, outputs of the two CNNs are concatenated with the original text to form the multi-modal embedding. This can be used as input to independent downstream tasks.}
    \label{pip}
\end{figure*}
Algorithm 1 provides the pseudo-code for the whole Interaction Canonical Correlation Network (ICCN).
\begin{algorithm}
\begin{algorithmic}
\renewcommand{\algorithmicrequire}{\textbf{Input:}}
\REQUIRE Data $H_{t}\in\mathbb{R}^{d_{t} \times N}$, $H_{a}\in \mathbb{R}^{d_{a} \times l_{a} \times N}$, $H_{v} \in \mathbb{R}^{d_{v} \times l_{v} \times N}$, $epoch$, $\eta$
%\STATE Initialize $H_{tv} = H_{t} \otimes H_{v}$, $H_{ta} = H_{t} \otimes H_{a}$\\
\STATE Initialize \textbf{$W_{a}$} of ($ \textrm{CNN1D}_{a}, \textrm{LSTM}_{a}$, and $\textrm{CNN2D}_{ta}$)
\STATE Initialize \textbf{$W_{v}$} of ($ \textrm{CNN1D}_{v}, \textrm{LSTM}_{v}$, and $\textrm{CNN2D}_{tv}$)
%\STATE Initialize $LSTM_{a}$, $LSTM_{v}$
%\STATE Initialize $CNN2d_{tv}$, $CNN2d_{ta}$
\WHILE{$epoch > 0$}
\STATE $H_{a1} = \textrm{CNN1D}_{a} (H_{a})$
\STATE $H_{v1} = \textrm{CNN1D}_{v} (H_{v})$
\STATE $H_{a2} = \textrm{LSTM}{a} (H_{a1})$
\STATE $H_{v2} = \textrm{LSTM}{v} (H_{v1})$
\STATE $H_{tv} = H_{t} \otimes H_{v2}$, $H_{ta} = H_{t} \otimes H_{a2}$
\STATE $K_{tv} =\textrm{CNN2D}_{tv} (H_{tv})$
\STATE $K_{ta} =\textrm{CNN2D}_{ta} (H_{ta})$ 
\STATE Compute gradients $\nabla W_{v}$, $\nabla W_{a}$ of:

\STATE $\min\limits_{W_{v},W_{a}}\textrm{CCALoss}(K_{tv}, K_{ta})$
%\STATE $CCALoss = $ $CCA(R_{tv}, R_{ta})$.
\STATE Update:
\STATE $W_{v} \leftarrow W_{v} - \eta \nabla W_{v}$
\STATE $W_{a} \leftarrow W_{a} - \eta \nabla W_{a}$
\STATE $epoch \leftarrow epoch - 1$
\ENDWHILE
\renewcommand{\algorithmicensure}{\textbf{Output:}}
\ENSURE $K_{tv}, K_{ta}$
\end{algorithmic}

\caption{Interaction Canonical Correlation Network}
\end{algorithm}
\label{ICC}
\subsection{Pipe-line for Downstream Tasks}
The ICCN method acts as a feature extractor. In order to test its performance, an additional downstream classifier is also required. Uni-modal features can be extracted using a variety of simple extraction schemes as well as by learning features using more complex neural network based models such as a sequential LSTM. 
Once uni-modal features for text, video, and audio  have been obtained, the ICCN can be used to learn text-based audio feature $K_{ta}$ and text-based video feature $K_{tv}$. The final multi-modal embedding can be formed as the concatenation of the text-based audio, the original text, and the text-based video features, which are denoted as $[ K_{ta}; H_{t}; K_{tv}]$. This $[K_{ta}; H_{t}; K_{tv}]$ can then be used as an input to downstream classifiers such as logistic regression or multilayer perceptron.
Figure~\ref{pip} shows the pipe-line using the ICCN for downstream tasks in our work.

\section{Experiment Settings}\label{exps}
This section describes the experimental datasets and baseline algorithms against which ICCN is compared.
% This section describes several multi-modal sentiment analysis experiments that analyze the usefulness of the multi-modal features learned from the ICCN.
\subsection{Datasets}
The proposed algorithm is tested using three public benchmark multi-modal sentiment analysis and emotion recognition datasets: CMU-MOSI \cite{zadeh2016mosi}, CMU-MOSEI \cite{zadeh2018multimodal}, and IEMOCAP\cite{busso2008iemocap}. 
Both CMU-MOSI and CMU-MOSEI's raw features, and most of the corresponding extracted features can be acquired from the CMU-MultimodalSDK \cite{zadeh2018multi}. 
\begin{itemize}
\item \textbf{CMU-MOSI:} This dataset is a multi-modal dataset built on 93 Youtube movie reviews. Videos are segmented to 2198 utterance clips, and each utterance example is annotated on a scale of [-3, 3] to reflect sentiment intensity. -3 means an extremely negative and 3 means an extremely positive sentiment. This data set is divided into three parts, training (1283 samples), validation (229 samples) and test (686 samples).

\item \textbf{CMU-MOSEI:} This dataset is similar to the CMU-MOSI, but is larger in size. It consists of 22856 annotated utterances extracted from Youtube videos. Each utterance can be treated as an individual multi-modal example. Train, validation, and test sets contain 16326, 1871, and 4659 samples respectively.

\item \textbf{IEMOCAP:} This dataset contains 302 videos in which speakers performed 9 different emotions (angry, excited, fear, sad, surprised, frustrated, happy, disappointed and neutral). Those videos are divided into short segments with emotion annotations. Due to the imbalance of some emotion labels, we follow experiments in previous papers \cite{wang2018words,liu2018efficient} where only four emotions (angry, sad, happy, and neutral) are used to test the performance of the algorithm. Train, validation, and test partitions contain 2717, 789, and 938 data samples respectively.
\end{itemize}
\subsection{Multi-modal Features}
The following uni-modal features are used prior to combinations,
\begin{itemize}
\item \textbf{Text Features:} For MOSI and MOSEI, we use a pre-trained transformer model BERT~\cite{devlin2018bert} to extract utterance level text features, (many other approaches use Glove word-level embeddings followed by a LSTM). The motivation behinds using BERT is 1) BERT is the state-of-the-art in sentence encoding algorithms and has demonstrated tremendous success in several downstream text applications such as sentiment analysis, question-answering, semantic similarity tasks etc, 2) using BERT simplifies the training pipe-line, with a large focus now towards improving the performance of ICCN on a particular downstream task. We input the raw text to the pre-trained uncased BERT-Base model (without fine-tuning). Sentence encodings output from BERT are used as the text features. Each individual text feature is of size 768. For IEMOCAP, we used InferSent\cite{conneau2017supervised} to encode utterance level text. Since data is provided in the form of word indices for GLOVE embeddings rather than raw text, we use InferSent; a BiLSTM layer followed by a max-pooling layer to learn sentence embeddings.
\item \textbf{Audio Features:} The audio feature is extracted by using COVAREP~\cite{degottex2014covarep}, which is a public software used for extracting acoustic features such as pitch, volume, and frequency. The CMU-MultimodalSDK provides COVAREP feature sequence for every multi-modal example, the dimension of each frame's audio feature is 74.
\item \textbf{Video Features:} Facet\cite{iMo} has been used for extracting facial expression features such as action units and face pose. Similarly, every multimodal example's video feature sequence is also obtained from the CMU-MultimodalSDK. The size of each frame's video feature is 35.
\end{itemize}

\subsection{Baseline Methods}
We consider a variety of baseline methods for multi-modal embedding comparison. In order to focus on the multi-modal embedding itself, we input each multi-modal embedding to the same downstream task classifier (or regressor).
Experimental comparisons are reported in two parts 1) we report the effectiveness of DCCA over the simpler CCA based methods when used as inputs to the ICCN and 2) we compare ICCN against newer utterance level embeddings algorithms that learn features for a down stream task in an end-to-end fashion.
The following baselines are used in our comparisons,
\begin{itemize}
\item \textbf{Uni-modal and Concatenation}: This is the simplest baseline in which uni-modal features are concatenated to obtain a multi-modal embedding.

\item \textbf{Linear CCA:} CCA~\cite{10.1093/biomet/28.3-4.321} considers linear transformations for different inputs. We use the CCA to learn a new common space for audio and video modes, and combine the learned audio and video features with the original text embedding. This is because, ~\cite{sun2019multi} showed that using a CCA-based method to correlate audio and video is more effective that correlating audio-text or video-text.

\item \textbf{Kernel-CCA:} Kernel-CCA~\cite{akaho2006kernel} introduces a nonlinearity via kernel maps. Kernel-CCA can be used exactly like CCA.

\item \textbf{GCCA:} Generalized CCA~\cite{tenenhaus2011regularized} learns a common subspace across multiple views. We use GCCA in two ways: 1) use the GCCA output embedding directly and 2) combine the GCCA output embedding with the original text embedding.
 
\item \textbf{DCCA:} A Deep CCA based algorithm proposed by~\cite{sun2019multi}. Audio and video features are simply concatenated and then be correlated with text features using DCCA. Outputs of the DCCA are again concatenated with raw text, audio, and video features to formulate the multimodal embedding.
\end{itemize}
In the proposed ICCN algorithm, text features are encoded by a pre-trained BERT transformer or by InferSent. This is unlike most of the state-of-the-art algorithms that obtain sentence level encodings by passing word embeddings through variants of RNNs. However, since the idea is to compare modal features, we also choose the following three state-of-the-art utterance-level (i.e. sentence-level) fusion models (whose core algorithm is agnostic to the text encoding architecture) as additional baselines. To make the comparison fair, these methods use the same features as ours.
\begin{itemize} 
\item \textbf{TFN:} Tensor Fusion Network (TFN)~\cite{zadeh2017tensor} combines individual modal's embeddings via calculating three different outer-product sub-tensors: unimodal, bimodal, and trimodal. All tensors will then be flattened and used as a multi-modal embedding vector.

\item \textbf{LMF:} Low-Rank Multimodal Fusion (LMF)~\cite{liu2018efficient} learns the multimodal embedding based the similar tensor processing of TFN, but with an additional low-rank factor for reducing computation memory.

\item \textbf{MFM:} Multimodal Factorization Model (MFM)~\cite{tsai2018learning} is consists of a discriminative model for prediction and a generative model for reconstructing input data. A comprehensive multimodal embedding is learned via optimizing the generative-discriminative objective simultaneously. 
\end{itemize}

\begin{table*}
\centering
\resizebox{1.5\columnwidth}{!}{
\begin{tabular}{|l|l|l|l|l|l|l|l|l|l|l|}
\hline
\multicolumn{1}{|c|}{Data View}&\multicolumn{5}{|c|}{CMU-MOSI}&\multicolumn{5}{|c|}{CMU-MOSEI}\\
\cline{2-11}&Acc-2&F-score&MAE&Acc-7&Corr&Acc-2&F-score&MAE&Acc-7&Corr\\
\hline
Audio&45.15&45.83&1.430&16.21&0.248&58.75&59.23&0.785&38.59&0.298\\
Video&48.10&49.06&1.456&15.51&0.339&59.25&59.90&0.770&36.09&0.288\\
Text&80.80&80.17&0.897&35.92&0.688&82.83&83.02&0.582&48.76&0.681\\
Text+Video&81.00&80.91&0.920&35.11&0.676&82.86&83.01&0.581&47.92&0.674\\
Text+Audio&80.59&80.56&0.909&35.08&0.672&82.80&82.96&0.582&49.02&0.689\\
Audio+Video+Text&80.94&81.00&0.895&36.41&0.689&82.72&82.87&0.583&50.11&0.692\\
\hline
CCA&79.45&79.35&0.893&34.15&0.690&82.94&83.06&0.573&50.23&0.690\\
KCCA&79.82&79.91&0.889&34.76&0.689&83.05&83.14&0.574&50.09&0.692\\
GCCA&62.50&62.15&1.403&17.29&0.533&75.12&75.46&0.653&45.33&0.602\\
GCCA+Text&77.80&77.87&1.107&25.94&0.658&82.75&82.90&0.613&46.06&0.644\\
DCCA&80.60$^{*}$&80.57$^{*}$&0.874&35.51&0.703&83.62$^{*}$&83.75$^{*}$&0.579&50.12&0.707\\
\hline
TFN&80.82&80.77&0.901&34.94&0.698&82.57&82.09&0.593&50.21&0.700\\
LMF&82.53&82.47&0.917&33.23&0.695&82.03&82.18&0.623&48.02&0.677\\
MFM&81.72&81.64&0.877&35.42&0.706&\textbf{84.40}&\textbf{84.36}&0.568&51.37&\textbf{0.717}\\
\hline
\textbf{ICCN}&\textbf{83.07}&\textbf{83.02}&\textbf{0.862}&\textbf{39.01}&\textbf{0.714}&84.18&84.15&\textbf{0.565} &\textbf{51.58}&0.713\\
\hline

\end{tabular}
}
\caption{Results for experiments on CMU-MOSI and CMU-MOSEI. Best numbers are in bold. For accuracy, F-score, and Correlation, higher is better. For mean absolute error, lower is better. Results marked with $*$ are reported in original papers. For TFN, LMF, and MFM, we re-did experiments with using our features for a fair comparison.}
\end{table*}

\subsection{Ablation studies of ICCN}
In order to analyze the usefulness of different components of the ICCN, we consider the following two questions: 

\textbf{Question 1}: Is using canonical correlation better than using other methods like Cosine-Similarity?

\textbf{Question 2}: Is learning the interactions between text and video (or audio) useful?

We design several experiments to address these two questions, First, we replace the CCA Loss with Cosine-Similarity Loss while leaving other parts of the ICCN unchanged. Second, instead of using the outer-product of audio (or video) and text as input to the CCA Loss, we use audio and video directly as the input to DCCA network.
We compare different ICCN variants' performance to prove the usefulness of each component of the network.

\subsection{Evaluation Methods}
To evaluate ICCN against previous baselines, the following performance metrics as described in~\cite{liu2018efficient,zadeh2017tensor,tsai2018learning} are reported,
\begin{itemize}

\item On the CMU-MOSI and CMU-MOSEI we report four performance metrics, i) binary accuracy, ii) F1-score iii) mean absolute error and iv) 7-class sentiment level / Correlation with human labeling.

\item On the IEMOCAP we used i) binary accuracy and ii) F1-score for evaluation.

\end{itemize}

\textbf{Evaluations Details on CMU-MOSI, CMU-MOSEI:} The original MOSI and MOSEI datasets are labeled in the range [-3,3]. The author of the datasets suggests a criterion for building binary labels: examples with label in [-3, 0) are considered to have negative sentiment while examples with label in (0, 3] are considered to have positive sentiment. 7-class sentiment level is also calculated based on the label distribution in [-3,3]. The correlation of predicted results with human labeling is also used as a criteria.  

\subsection{Hyperparameter Tuning}
A basic Grid-Search is used to tune hyperparameters, and the best hyperparameter settings for the ICCN are chosen according to its performance on the validation dataset. For ICCN, hyperparameters and tuning ranges are: learning rate ($1e-5$--$1e-3$), mini-batch size ($128$--$512$), the number of epoch ($10$--$100$), hidden dimensions of MLP ($64$--$512$), and output dimension of the CCA projection ($30$--$100$). ReLU is used as the activation function, RMSProp is used as the optimizer. 

Whenever the training of the ICCN with a specific hypyerparameter setting has finished, features learned from the ICCN are used as input to the same downstream task models (a simple MLP is used in this work). Test results are reported by using the best hyperparameter setting learned above.
\begin{table*}
\centering
\resizebox{1.4\columnwidth}{!}{
\begin{tabular}{|l|l|l|l|l|l|l|l|l|}
\hline
\multicolumn{1}{|c|}{Data View}&\multicolumn{8}{|c|}{IEMOCAP}\\
\hline
\multicolumn{1}{|c|}{Emotions}&\multicolumn{2}{|c|}{Happy}&\multicolumn{2}{|c|}{Angry}&\multicolumn{2}{|c|}{Sad}&\multicolumn{2}{|c|}{Neutral}\\
%\hline
\cline{2-9}&Acc-2&F-score&Acc-2&F-score&Acc-2&F-score&Acc-2&F-score\\
\hline
Audio&84.03&81.09&85.49&84.03&82.75&80.26&63.08&59.24\\
Video&83.14&80.36&85.91&83.27&81.19&80.35&62.30&58.19\\
Text&84.80&81.17&85.12&85.21&84.18&83.63&66.02&63.52\\
Text+Video&85.32&82.01&85.22&85.10&83.33&82.96&65.82&65.81\\
Text+Audio&85.10&83.47&86.09&84.99&83.90&83.91&66.96&65.02\\
Audio+Video+Text&86.00&83.37&86.37&85.88&84.02&83.71&66.87&65.93\\
\hline
CCA&85.91&83.32&86.17&84.39&84.19&83.71&67.22&64.84\\
KCCA&86.54&84.08&86.72&86.32&85.03&84.91&68.29&65.93\\
GCCA&81.15&80.33&82.47&78.06&83.22&81.75&65.34&59.99\\
GCCA+Text&87.02&83.44&88.01&88.00&84.79&83.26&68.26&67.61\\
DCCA&86.99&84.32&87.94&87.85&86.03&84.36&68.87&65.93\\
\hline
TFN&86.66&84.03&87.11&87.03&85.64&85.75&68.90&68.03\\
LMF&86.14&83.92&86.24&86.41&84.33&84.40&69.62&68.75\\
MFM&86.67&84.66&86.99&86.72&85.67&85.66&\textbf{70.26}&\textbf{69.98}\\
\hline
\textbf{ICCN}&\textbf{87.41}&\textbf{84.72}&\textbf{88.62}&\textbf{88.02}&\textbf{86.26}&\textbf{85.93}&69.73&68.47 \\
\hline
\end{tabular}
}
\caption{Results for experiments on IEMOCAP. Best numbers are in bold. For accuracy and F-score, higher is better. For TFN, LMF, and MFM, we re-did experiments with using our features for a fair comparison.}
\end{table*}

\begin{table}
\centering
\resizebox{\columnwidth}{!}{
\begin{tabular}{|l|l|l|l|l|l|l|l|l|l|l|}
\hline
\multicolumn{1}{|c|}{Data View}&\multicolumn{5}{|c|}{CMU-MOSI}&\multicolumn{5}{|c|}{CMU-MOSEI}\\
\cline{2-11}&Acc-2&F-score&MAE&Acc-7&Corr&Acc-2&F-score&MAE&Acc-7&Corr\\
\hline
\textbf{ICCN} & \textbf{83.07}& \textbf{83.02} &\textbf{0.862} &\textbf{39.01} &\textbf{0.714} &\textbf{84.18} &\textbf{84.15} &\textbf{0.565} &\textbf{51.58} &\textbf{0.713}\\
ICCN$_{1}$(no text)&82.13&82.05&0.874&35.51&0.703&83.01&83.10&0.575&50.12&0.707 \\
ICCN$_{2}$(cos)&82.32&82.27&0.876&36.01&0.702&82.98&82.90&0.575&50.63&0.700\\
ICCN$_{3}$(no text + cos)&81.49&81.58&0.889&35.77&0.692&82.59&82.73&0.578&50.21&0.696\\
\hline
\end{tabular}
}
\caption{Ablation studies of ICCN on CMU-MOSI and CMU-MOSEI. ICCN$_{1-3}$ denote different variants, ``no text" means applying DCCA to audio and video directly instead of applying to the outer-product with text; ``cos" means replacing CCA Loss with Cosine-Similarity Loss;``no text + cca" means removing outer-product with text and use Cosine-Similarity Loss.}
\end{table}

\section{Discussion of Empirical Results}\label{discu}
This section presents and discusses results on the CMU-MOSI, CMU-MOSEI, and IEMOCAP datasets.

\subsection{Performance on Benchmark Datasets}

Tables 1 and 2 present results of experiments on the CMU-MOSI, CMU-MOSEI, and IEMOCAP datasets. 
\begin{itemize}
\item First, when compared with results of using uni-modal and simple concatenations, ICCN outperforms all of them in all of the criteria. Note that the performance of the text feature is always better than the performance of the audio and video, and that a simple concatenation of text, video, and audio does not work well. This shows the advance of highly-developed pre-trained text features capable of improving the overall multimodal task performance. However, it also shows the challenge of how to effectively combine such a highly developed text feature with audio and video features.  
\item Second, ICCN also outperforms other CCA-based methods. The results of using other CCA-based methods show that they cannot improve the multimodal embedding's performance. We argue this occurs because 1) CCA / KCCA / GCCA do not exploit the power of neural network architectures so that their learning capacities are limited. 2) Using DCCA without learning the interactions between text-based audio and text-based video may sacrifice useful information.

\item Third, the ICCN still achieves better or similar results when compared with other neural network based state-of-the-art methods (TFN, LMF, and MFM). These results demonstrate the ICCN's competitive performance.
\end{itemize}

\subsection{Results of Ablation Studies}

Table 3 shows results of using variants of ICCN on CMU-MOSI and CMU-MOSEI datasets. 

First, using the CCA Loss performs better than using Cosine-Similarity Loss with or without the outer-product. This is reasonable as the DCCA is able to learn the hidden relationships (with the help of non-linear transformations) but cosine-similarity is restrained by the original coordinates. To further verify this, we also record changes of canonical correlation and cosine similarity between text-based audio and text-based video (i.e., between the two outputs of the CNNs in the ICCN) by using CCA Loss or Cosine Similarity Loss for the ICCN with using the CMU-MOSI dataset. Curves in Figures 2 and 3 summarize the results of the experiments. Results show that maximizing the canonical correlation by using the CCA Loss does not necessarily increase the cosine similarity, and vice versa. This demonstrates that the canonical correlation is a genuinely different objective function than cosine similarity, and explains the different behaviors in the downstream applications. CCA is capable of learning hidden relationships between inputs that the cosine similarity does not see.

Second, learning the interactions between non-text and text performs better than using audio and video directly. This also makes sense because audio and video are more correlated when they are based on the same text, thus learning text-based audio and text-based video performs better. In summary, Table 3 shows the usefulness of using a text-based outer-product together with DCCA.

 \begin{figure}[ht!]
     \centering
     \includegraphics[width=.75\columnwidth]{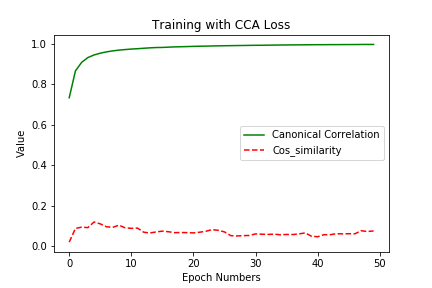}
     \caption{Changes of mean canonical correlation and mean cosine similarity between text-based audio and text-based video when training with CCA Loss: The network learns to maximize the canonical correlation but the cosine similarity isn't affected.}
     \label{cca_loss}
 \end{figure}
 \begin{figure}[ht!]
     \centering
     \includegraphics[width=.75\columnwidth]{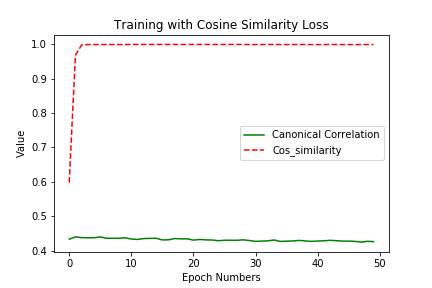}
     \caption{Changes of mean canonical correlation and cosine similarity between text-based audio and text-based video when training with Cosine Similarity Loss: convergence of cosine similarity doesn't affect canonical correlation.}
     \label{cos_loss}
 \end{figure}

\section{Conclusion and Future Work}\label{conclu}

This paper has proposed the ICCN method, which uses canonical correlation to analyze hidden relationships between text, audio, and video. Testing on a multi-modal sentiment analysis and emotion recognition task shows that the multi-modal features learned from the ICCN model can achieve state-of-the-art performance, and shows the effectiveness of the model. Ablation studies confirm the usefulness of different part of the network.

There is, of course, considerable room for improvement. Possible directions include learning dynamic intra-actions in each model together with inter-actions between different modes; learning the trade-off between maximum canonical correlation and best downstream task performance; and developing an interpretable end-to-end multi-modal canonical correlation model. In the future, we hope to move forward in the development of multi-modal machine learning.

\bibliographystyle{aaai} \bibliography{AAAI-SunZ.6643}

\begin{thebibliography}{}

\bibitem[\protect\citeauthoryear{Akaho}{2006}]{akaho2006kernel}
Akaho, S.
\newblock 2006.
\newblock A kernel method for canonical correlation analysis.
\newblock {\em arXiv preprint cs/0609071}.

\bibitem[\protect\citeauthoryear{Andrew \bgroup et al\mbox.\egroup
  }{2013}]{andrew2013deep}
Andrew, G.; Arora, R.; Bilmes, J.; and Livescu, K.
\newblock 2013.
\newblock Deep canonical correlation analysis.
\newblock In {\em ICML},  1247--1255.

\bibitem[\protect\citeauthoryear{Busso \bgroup et al\mbox.\egroup
  }{2008}]{busso2008iemocap}
Busso, C.; Bulut, M.; Lee, C.-C.; Kazemzadeh, A.; Mower, E.; Kim, S.; Chang,
  J.~N.; Lee, S.; and Narayanan, S.~S.
\newblock 2008.
\newblock Iemocap: Interactive emotional dyadic motion capture database.
\newblock {\em Language resources and evaluation} 42(4):335.

\bibitem[\protect\citeauthoryear{Chen \bgroup et al\mbox.\egroup
  }{2017}]{chen2017multimodal}
Chen, M.; Wang, S.; Liang, P.~P.; Baltru{\v{s}}aitis, T.; Zadeh, A.; and
  Morency, L.-P.
\newblock 2017.
\newblock Multimodal sentiment analysis with word-level fusion and
  reinforcement learning.
\newblock In {\em ICMI},  163--171.
\newblock ACM.

\bibitem[\protect\citeauthoryear{Conneau \bgroup et al\mbox.\egroup
  }{2017}]{conneau2017supervised}
Conneau, A.; Kiela, D.; Schwenk, H.; Barrault, L.; and Bordes, A.
\newblock 2017.
\newblock Supervised learning of universal sentence representations from
  natural language inference data.
\newblock {\em arXiv preprint arXiv:1705.02364}.

\bibitem[\protect\citeauthoryear{Degottex \bgroup et al\mbox.\egroup
  }{2014}]{degottex2014covarep}
Degottex, G.; Kane, J.; Drugman, T.; Raitio, T.; and Scherer, S.
\newblock 2014.
\newblock Covarep—a collaborative voice analysis repository for speech
  technologies.
\newblock In {\em ICASSP},  960--964.
\newblock IEEE.

\bibitem[\protect\citeauthoryear{Devlin \bgroup et al\mbox.\egroup
  }{2018}]{devlin2018bert}
Devlin, J.; Chang, M.-W.; Lee, K.; and Toutanova, K.
\newblock 2018.
\newblock Bert: Pre-training of deep bidirectional transformers for language
  understanding.
\newblock {\em arXiv preprint arXiv:1810.04805}.

\bibitem[\protect\citeauthoryear{Gers, Schmidhuber, and
  Cummins}{1999}]{gers1999learning}
Gers, F.~A.; Schmidhuber, J.; and Cummins, F.
\newblock 1999.
\newblock Learning to forget: Continual prediction with lstm.
\newblock {\em ICANN}.

\bibitem[\protect\citeauthoryear{Haq and Jackson}{2011}]{haq2011multimodal}
Haq, S., and Jackson, P.~J.
\newblock 2011.
\newblock Multimodal emotion recognition.
\newblock In {\em MAPAS}. IGI Global.
\newblock  398--423.

\bibitem[\protect\citeauthoryear{Hazarika \bgroup et al\mbox.\egroup
  }{2018}]{hazarika2018cascade}
Hazarika, D.; Poria, S.; Gorantla, S.; Cambria, E.; Zimmermann, R.; and
  Mihalcea, R.
\newblock 2018.
\newblock Cascade: Contextual sarcasm detection in online discussion forums.
\newblock {\em arXiv preprint arXiv:1805.06413}.

\bibitem[\protect\citeauthoryear{Hotelling}{1936}]{10.1093/biomet/28.3-4.321}
Hotelling, H.
\newblock 1936.
\newblock {Relations between two set of variables}.
\newblock {\em Biometrika} 28(3-4):321--377.

\bibitem[\protect\citeauthoryear{iMotions}{2017}]{iMo}
iMotions.
\newblock 2017.
\newblock Facial expression analysis.
\newblock {\em imotions.com}.

\bibitem[\protect\citeauthoryear{Lawrence \bgroup et al\mbox.\egroup
  }{1997}]{lawrence1997face}
Lawrence, S.; Giles, C.~L.; Tsoi, A.~C.; and Back, A.~D.
\newblock 1997.
\newblock Face recognition: A convolutional neural-network approach.
\newblock {\em IEEE transactions on neural networks} 8(1):98--113.

\bibitem[\protect\citeauthoryear{Lee \bgroup et al\mbox.\egroup
  }{2014}]{lee2014supervised}
Lee, G.; Singanamalli, A.; Wang, H.; Feldman, M.~D.; Master, S.~R.; Shih,
  N.~N.; Spangler, E.; Rebbeck, T.; Tomaszewski, J.~E.; and Madabhushi, A.
\newblock 2014.
\newblock Supervised multi-view canonical correlation analysis (smvcca):
  Integrating histologic and proteomic features for predicting recurrent
  prostate cancer.
\newblock {\em IEEE transactions on medical imaging} 34(1):284--297.

\bibitem[\protect\citeauthoryear{Liang \bgroup et al\mbox.\egroup
  }{2018}]{liang2018multimodal}
Liang, P.~P.; Liu, Z.; Zadeh, A.; and Morency, L.-P.
\newblock 2018.
\newblock Multimodal language analysis with recurrent multistage fusion.
\newblock {\em arXiv preprint arXiv:1808.03920}.

\bibitem[\protect\citeauthoryear{Liu \bgroup et al\mbox.\egroup
  }{2018}]{liu2018efficient}
Liu, Z.; Shen, Y.; Lakshminarasimhan, V.~B.; Liang, P.~P.; Zadeh, A.; and
  Morency, L.-P.
\newblock 2018.
\newblock Efficient low-rank multimodal fusion with modality-specific factors.
\newblock {\em ACL}.

\bibitem[\protect\citeauthoryear{Liu \bgroup et al\mbox.\egroup
  }{2019}]{liu2019}
Liu, F.; Guan, B.; Zhou, Z.; Samsonov, A.; G~Rosas, H.; Lian, K.; Sharma, R.;
  Kanarek, A.; Kim, J.; Guermazi, A.; and Kijowski, R.
\newblock 2019.
\newblock Fully-automated diagnosis of anterior cruciate ligament tears on knee
  mr images using deep learning.
\newblock {\em Radiology} 1.

\bibitem[\protect\citeauthoryear{Martin and
  Maes}{1979}]{martin1979multivariate}
Martin, N., and Maes, H.
\newblock 1979.
\newblock {\em Multivariate analysis}.
\newblock Academic press London.

\bibitem[\protect\citeauthoryear{Morency, Mihalcea, and
  Doshi}{2011}]{morency_towards_2011}
Morency, L.-P.; Mihalcea, R.; and Doshi, P.
\newblock 2011.
\newblock Towards {Multimodal} {Sentiment} {Analysis}: {Harvesting} {Opinions}
  from {The} {Web}.
\newblock In {\em {ICMI} 2011}.

\bibitem[\protect\citeauthoryear{Poria \bgroup et al\mbox.\egroup
  }{2016}]{poria2016convolutional}
Poria, S.; Chaturvedi, I.; Cambria, E.; and Hussain, A.
\newblock 2016.
\newblock Convolutional mkl based multimodal emotion recognition and sentiment
  analysis.
\newblock In {\em ICDM},  439--448.
\newblock IEEE.

\bibitem[\protect\citeauthoryear{Rotman, Vuli{\'c}, and
  Reichart}{2018}]{rotman2018bridging}
Rotman, G.; Vuli{\'c}, I.; and Reichart, R.
\newblock 2018.
\newblock Bridging languages through images with deep partial canonical
  correlation analysis.
\newblock In {\em ACL(Volume 1: Long Papers)},  910--921.

\bibitem[\protect\citeauthoryear{Sarma, Liang, and
  Sethares}{2018}]{sarma2018domain}
Sarma, P.~K.; Liang, Y.; and Sethares, W.~A.
\newblock 2018.
\newblock Domain adapted word embeddings for improved sentiment classification.
\newblock {\em arXiv preprint arXiv:1805.04576}.

\bibitem[\protect\citeauthoryear{Soleymani \bgroup et al\mbox.\egroup
  }{2017}]{soleymani2017survey}
Soleymani, M.; Garcia, D.; Jou, B.; Schuller, B.; Chang, S.-F.; and Pantic, M.
\newblock 2017.
\newblock A survey of multimodal sentiment analysis.
\newblock {\em Image and Vision Computing} 65:3--14.

\bibitem[\protect\citeauthoryear{Sun \bgroup et al\mbox.\egroup
  }{2019}]{sun2019multi}
Sun, Z.; Sarma, P.~K.; Sethares, W.; and Bucy, E.~P.
\newblock 2019.
\newblock Multi-modal sentiment analysis using deep canonical correlation
  analysis.
\newblock {\em arXiv preprint arXiv:1907.08696}.

\bibitem[\protect\citeauthoryear{Tenenhaus and
  Tenenhaus}{2011}]{tenenhaus2011regularized}
Tenenhaus, A., and Tenenhaus, M.
\newblock 2011.
\newblock Regularized generalized canonical correlation analysis.
\newblock {\em Psychometrika} 76(2):257.

\bibitem[\protect\citeauthoryear{Tsai \bgroup et al\mbox.\egroup
  }{2018}]{tsai2018learning}
Tsai, Y.-H.~H.; Liang, P.~P.; Zadeh, A.; Morency, L.-P.; and Salakhutdinov, R.
\newblock 2018.
\newblock Learning factorized multimodal representations.
\newblock {\em arXiv preprint arXiv:1806.06176}.

\bibitem[\protect\citeauthoryear{Tsai \bgroup et al\mbox.\egroup
  }{2019}]{tsai2019multimodal}
Tsai, Y.-H.~H.; Bai, S.; Liang, P.~P.; Kolter, J.~Z.; Morency, L.-P.; and
  Salakhutdinov, R.
\newblock 2019.
\newblock Multimodal transformer for unaligned multimodal language sequences.
\newblock {\em arXiv preprint arXiv:1906.00295}.

\bibitem[\protect\citeauthoryear{Wang \bgroup et al\mbox.\egroup
  }{2018}]{wang2018words}
Wang, Y.; Shen, Y.; Liu, Z.; Liang, P.~P.; Zadeh, A.; and Morency, L.-P.
\newblock 2018.
\newblock Words can shift: Dynamically adjusting word representations using
  nonverbal behaviors.
\newblock {\em arXiv preprint arXiv:1811.09362}.

\bibitem[\protect\citeauthoryear{Wang, Li, and
  Lazebnik}{2016}]{wang2016learning}
Wang, L.; Li, Y.; and Lazebnik, S.
\newblock 2016.
\newblock Learning deep structure-preserving image-text embeddings.
\newblock In {\em CVPR},  5005--5013.

\bibitem[\protect\citeauthoryear{Zadeh \bgroup et al\mbox.\egroup
  }{2016}]{zadeh2016mosi}
Zadeh, A.; Zellers, R.; Pincus, E.; and Morency, L.-P.
\newblock 2016.
\newblock Mosi: multimodal corpus of sentiment intensity and subjectivity
  analysis in online opinion videos.
\newblock {\em arXiv preprint arXiv:1606.06259}.

\bibitem[\protect\citeauthoryear{Zadeh \bgroup et al\mbox.\egroup
  }{2017}]{zadeh2017tensor}
Zadeh, A.; Chen, M.; Poria, S.; Cambria, E.; and Morency, L.-P.
\newblock 2017.
\newblock Tensor fusion network for multimodal sentiment analysis.
\newblock {\em arXiv preprint arXiv:1707.07250}.

\bibitem[\protect\citeauthoryear{Zadeh \bgroup et al\mbox.\egroup
  }{2018a}]{zadeh2018memory}
Zadeh, A.; Liang, P.~P.; Mazumder, N.; Poria, S.; Cambria, E.; and Morency,
  L.-P.
\newblock 2018a.
\newblock Memory fusion network for multi-view sequential learning.
\newblock In {\em AAAI}.

\bibitem[\protect\citeauthoryear{Zadeh \bgroup et al\mbox.\egroup
  }{2018b}]{zadeh2018multi}
Zadeh, A.; Liang, P.~P.; Poria, S.; Vij, P.; Cambria, E.; and Morency, L.-P.
\newblock 2018b.
\newblock Multi-attention recurrent network for human communication
  comprehension.
\newblock In {\em AAAI}.

\bibitem[\protect\citeauthoryear{Zadeh \bgroup et al\mbox.\egroup
  }{2018c}]{zadeh2018multimodal}
Zadeh, A.~B.; Liang, P.~P.; Poria, S.; Cambria, E.; and Morency, L.-P.
\newblock 2018c.
\newblock Multimodal language analysis in the wild: Cmu-mosei dataset and
  interpretable dynamic fusion graph.
\newblock In {\em ACL},  2236--2246.

\end{thebibliography}
\end{document}